\documentclass[pmlr]{jmlr}

\RequirePackage{graphicx}
\usepackage{booktabs}
\usepackage{subcaption}
\usepackage{longtable}
\usepackage{ragged2e}
\usepackage{caption} 
\makeatletter
\def\set@curr@file#1{\def\@curr@file{#1}} 
\makeatother
\usepackage[load-configurations=version-1]{siunitx} 

\usepackage{algorithmicx}

\usepackage{graphicx}

\usepackage{amsmath}
\usepackage{booktabs}
\usepackage{algorithm} 
\usepackage{float}
\usepackage{multicol}
\usepackage[normalem]{ulem}
\usepackage{mathtools, nccmath} 

\theorembodyfont{\upshape}
\theoremheaderfont{\scshape}
\theorempostheader{:}
\theoremsep{\newline}

\jmlryear{2024}
\jmlrworkshop{Machine Learning for Healthcare}
\jmlrvolume{252}

\title[G-Transformer]{G-Transformer: Counterfactual Outcome Prediction under Dynamic and Time-varying Treatment Regimes }

\author{\Name{Hong Xiong$^{1}$}\thanks{Indicates co-first authors.}
       \Email{hongxiong@hsph.harvard.edu}
       \AND
       \Name{Feng Wu$^{2}$}\footnotemark[1]
       \Email{wufeng@mit.edu}
       \AND  \Name{Leon Deng$^{2}$}
      \Email{lydeng@mit.edu} 
      \AND \Name{Megan Su$^{2}$}
      \Email{megansu@mit.edu}  
    \AND \Name{Zach Shahn$^{3,4}$}
      \Email{Zachary.Shahn@sph.cuny.edu}
    \AND \Name{Li{-}wei H Lehman$^{2}$}\thanks{Corresponding author.}
      \Email{lilehman@mit.edu}\\
            \addr 
       1. Harvard University, 
       Cambridge, MA, USA \\
       \addr 
      2. Massachusetts Institute of Technology,  Cambridge, MA, USA \\
      \addr
      3. City University of New York, NY, USA \\
      \addr
      4. MIT-IBM Watson AI Lab, Cambridge, MA
} 

\begin{document}

\maketitle

\begin{abstract}
 
  In the context of medical decision making, counterfactual prediction enables clinicians to predict treatment outcomes of interest under alternative courses of therapeutic actions given observed patient history. 
  In this work, we present G-Transformer for counterfactual outcome prediction under dynamic and time-varying treatment strategies.   Our approach leverages a Transformer architecture to capture complex, long-range dependencies in time-varying covariates while enabling g-computation, a causal inference method for estimating the effects of dynamic treatment regimes.
  Specifically, we use a Transformer-based encoder architecture to estimate the conditional distribution of relevant covariates given covariate and treatment history at each time point, then produces Monte Carlo estimates of counterfactual outcomes by simulating forward patient trajectories under treatment strategies of interest.  We evaluate G-Transformer extensively using two simulated longitudinal datasets from mechanistic models, and a real-world sepsis ICU dataset from MIMIC-IV. G-Transformer outperforms both classical and state-of-the-art counterfactual prediction models in these settings.   To the best of our knowledge, this is the first Transformer-based architecture that supports g-computation for counterfactual outcome prediction under dynamic and time-varying treatment strategies. 



\end{abstract}

\section{Introduction}

Clinicians often have to choose among multiple treatment options for their patients but do not have the ability to test every strategy before making a decision. In selecting among competing dynamic treatment strategies, it is useful to obtain counterfactual predictions regarding a patient's probability of experiencing adverse outcomes under each alternative strategy, based on their observed covariate history.

Counterfactual prediction in medical decision-making involves the estimation of potential future trajectories of covariates of interest  under alternative courses of action given observed history. Treatment strategies of interest are typically \textit{time-varying}, indicating decisions span multiple time points, and \textit{dynamic}, implying that each treatment decision at a given time point is influenced by the preceding history up to that point.
Recent works presented deep learning approaches to estimate time-varying treatment effects \citep{Lim2018,bica2020crn,bica2020TSD,melnychuk2022causal}.  However, most previous approaches focus on estimating counterfactual outcomes under static time-varying treatment strategies where treatments are not dependent on past covariate history. 
  
In this work, we present G-Transformer, a transformer-based \citep{vaswani2017attention} framework for counterfactual prediction under \textit{dynamic} and \textit{time-varying} treatment strategies.  G-Transformer supports g-computation, a causal inference technique for estimating treatment effects under dynamic treatment regimes \citep{robins1986,robins1987}.   
G-Transformer estimates the conditional distribution of relevant covariates given covariate and treatment history at each time point using an encoder architecture, then produces Monte Carlo estimates of counterfactual outcomes by simulating forward patient trajectories under treatment strategies of interest.  We evaluated G-Transformer extensively using multiple simulated datasets from two mechanistic models where ground-truths under the counterfactual strategies can be measured, and a real-world sepsis dataset to assess G-Transformer's potential clinical utility in counterfactual predictions under alternative dynamic fluid administration regimes.
Our contributions are the following: 
\begin{itemize} 
\item G-Transformer architecture. We introduce a novel Transformer-based architecture that enables counterfactual predictions under dynamic and time-varying treatment strategies to provide estimates of individual treatment effects. We present a custom sequential training procedure for supporting g-computation that performs better than the seq2seq training typically used in Transformers.

\item Evaluation under static time-varying treatment regimes. Using a simulated tumor growth dataset, we demonstrated that G-Transformer out-performed other state-of-the-art deep learning approaches, including rMSN \citep{Lim2018}, CRN \citep{bica2020crn}, G-Net \citep{GNet2021}, and Causal Transformer \citep{melnychuk2022causal}, in counterfactual prediction under static time-varying treatments, in which treatments are time-varying but do not depend on past covariate history. 

\item Evaluation under \textit{dynamic} and \textit{time-varying} treatment regimes.  We used CVSim \citep{CVSim2010}, a mechanistic model of the cardiovascular system, to simulate counterfactual patient trajectories under various \textit{dynamic} fluid and vasopressor administration strategies, and demonstrated that G-Transformer out-performed other baselines, including G-Net and a linear implementation of g-computation, in counterfactual prediction under dynamic and time-varying treatment regimes. 
\item Evaluation under real-world ICU dataset. Using real-world data of a sepsis cohort from the MIMIC-IV database \citep{mimic-iv,GoldbergerPhysioNet2000}, we evaluated the performance of G-Transformer in predicting outcomes under observational treatment regimes, and 
demonstrated G-Transformer's potential clinical utility in generating counterfactual predictions under alternative dynamic fluids administration regimes.
\end{itemize}

\subsection*{Generalizable Insights about Machine Learning in the Context of Healthcare}
 


 G-Transformer can facilitate treatment decision making by predicting time-varying treatment outcomes over time under alternative dynamic treatment strategies of interest.  Our extensive experiments show strong empirical performance of G-Transformer under both static and dynamic time-varying treatment strategies over other state-of-the-art techniques. 
Although we focus on applications in healthcare in this proposed work, the G-Transformer architecture can be applied to other sequential decision making tasks that involve dynamic and time-varying interventions.

\section{Related Work}
 
Recent works by \citet{Lim2018,bica2020crn,bica2020TSD} presented deep learning approaches to estimate time-varying treatment effects. \citet{bica2020crn} applied ideas from domain adaptation to estimate treatment effects over time while  \citet{Lim2018} used RNN regression models with history adjusted marginal structural models (MSMs)~\citep{ha_msm} to make counterfactual predictions.    
Causal Transformer (CT) has recently been proposed a Transformer-based architecture to estimate counterfactual outcomes \citep{melnychuk2022causal}. However, previous approaches focus on estimating counterfactual outcomes under time-varying treatment strategies where treatments are not dependent on past covariate history. None of these  approaches are designed to estimate effects under \emph{dynamic} treatment strategies, in which treatment depends on recent covariate history.  In contrast, our work focus on estimating treatment effects under \emph{dynamic} treatment strategies, in which treatments depend on covariate history. 

Dynamic-MSMs can be used to estimate expected counterfactual outcomes under restricted classes of simple dynamic treatment regimes \citep{dynMSM2010,ShahnEtAl2020}. It performs time-varying confounding adjustment via inverse probability weighting. The limitation with dynamic-MSMs is that the optimal treatment strategy of realistic complexity may not be included in the restricted class considered.  

G-computation can be used to estimate the average effect of a dynamic treatment regime (DTR) on the population, or the conditional effect given observed patient history. 
Previous implementations of g-computation used classic linear models \citep{taubman2009} or recurrent neural networks based architectures, such as LSTM \citep{GNet2021}.   G-Net \citep{GNet2021} is based on g-computation and used LSTMs for counterfactual prediction of time-varying treatment outcomes under alternative dynamic treatment strategies. In contrast, our work focused on a Transformer-based architecture to better capture complex dependencies in time-varying data. 
DeepACE \citep{frauen2023estimating} is a recurrent neural network based model that leverages the g-computation framework to address time-varying confounding, but primarily focuses on estimating the Average Causal Effect (ACE). 

\citet{Schulam2017}  proposes an implementation of continuous-time g-computation, focusing on static, time-varying treatment strategies. While their approach could potentially be extended to accommodate dynamic strategies, it is known that Gaussian Processes (GPs) become intractable for large datasets.
Recent work by \cite{WuEtAl2023} presented an alternating sequential model that used Transformers for clinical treatment outcome prediction, but their work did not support counterfactual prediction.  \cite{pmlr-v162-seedat22b} and \cite{holt2024ode} focused on Continuous-Time Modeling, attempting to predict counterfactual treatment effects over irregular time lengths. They utilized neural controlled differential equations to model patient historical trajectories and attempted to make uncertainty estimates based on this model. 
T4 \citep{liu2023estimating}  introduced a LSTM-based neural network architecture for predicting the counterfactual treatment effects of sepsis. They also estimated the uncertainty associated with treatment effects. 


\section{Methods}

\subsection{Background: G-computation and Problem Setup}

G-computation  \cite{robins1986} can be used to estimate the average effect of a dynamic treatment regime (DTR) on the population, or the conditional effect given observed patient history. In the latter case, we can estimate the expected counterfactual trajectory of outcomes of interest under alternative treatment strategies of interest for a particular patient history. 

\paragraph{Problem Setup} Our goal is to predict patient outcomes under alternative future treatment strategies given observed patient histories. We follow notations introduced in \cite{GNet2021} in this study.

\noindent Let:
\begin{itemize}
\item $t\in \{0,\ldots ,K\}$ denote time, assumed discrete, with $K$ being the end of followup;
\item $A_{t}$ denote the observed treatment action at time $t$;
\item $Y_t$ denote the observed outcome at time $t$ 
\item $L_{t}$ denote a vector of covariates at time $t$ that may influence treatment decisions or be associated with the outcome; 
\item $\bar{X}_{t}$ denote the history $X_{0},\ldots ,X_{t}$ and $\underline{X}_{t}$
denote the future $X_{t},\ldots ,X_{K}$ for arbitrary time varying variable~$X$.
\end{itemize}

\noindent 

At each time point, we assume the causal ordering $(L_t,A_t,Y_t)$, i.e. the treatments are administered after having observed the covariates at time step $t$. Let $H_t\equiv (\bar{L}_t,\bar{A}_{t-1})$ denote patient history preceding treatment at time $t$. A dynamic treatment strategy $g$ is a collection of functions $\{g_0,\ldots,g_K\}$, one per time point, such that $g_t$ maps $H_t$ onto a treatment action at time $t$. An example dynamic strategy might be to administer treatment A if the patient's vital signs, e.g. arterial blood pressure, drops below certain threshold.


Let $Y_t(g)$ denote the counterfactual outcome that would be observed at time $t$, had treatment strategy $g$ been followed from baseline \citep{robins1986}. Further, let $Y_t(\bar{A}_{m-1},\underline{g}_m)$ with $t\geq m$ denote the counterfactual outcome that would be observed had the patient received their observed treatments $\bar{A}_{m-1}$ through time $m-1$ then followed strategy $g$ from time $m$ onward. Here, the treatment strategy $g$ is typically specified by the domain experts, e.g. clinicians.  In this work, the outcome $Y_t$ can be deemed to be a variable in the vector $L_{t+1}$.


In counterfactual point prediction, our goal is to estimate expected counterfactual patient outcome trajectories 
\begin{align}\label{target}
\{E[Y_t(\bar{A}_{m-1},\underline{g}_m)|H_m], t\geq m\}
\end{align}
given observed patient history through time $m$ for any $m$ and any specified treatment strategy $g$, where $g$ is specified by a domain expert, e.g. a clinician. It may be of interest to estimate the counterfactual outcome distributions at future time points
\begin{align}\label{target2}
\{p(Y_t(\bar{A}_{m-1},\underline{g}_m)|H_m), t\geq m\}.
\end{align}

The g-computation algorithm enables the estimation of (\ref{target}) and (\ref{target2}) under the following assumptions \citep{robins1986}: Consistency, Sequential Exchangeability, and Positivity.  

\textbf{Assumption 3.1.} \textbf{Consistency} asserts that the observed outcome is equal to the counterfactual outcome that would have occurred had the observed treatment been applied. 

\textbf{Assumption 3.2.} \textbf{Sequential exchangeability}  requires that there is no unobserved confounding, i.e., all drivers of treatment decisions prognostic for the outcome are included in the data. 

\textbf{Assumption 3.3.} \textbf{Positivity} posits that the counterfactual treatment strategy of interest has some non-zero probability of actually being implemented. 


Under assumptions 3.1-3.3, for $t=m$ we have that 
{\small
\begin{align}
    p(Y_m(\bar{A}_{m-1},g_m)|H_m) = p(Y_m|H_m,A_m=g_m(H_m)),
\end{align}
}%
i.e. the conditional distribution of the counterfactual is the conditional distribution of the observed outcome given patient history and given that treatment follows the strategy of interest. For $t>m$,  we need to adjust for time-varying confounding. With $X_{i:j} = X_i,\ldots,X_j$ for any random variable X, under assumptions 1-3 the g-formula yields
{\small
\begin{align}\label{g_dist}
& p(Y_t(\bar{A}_{m-1},\underline{g}_m)=y|H_m) \nonumber \\
& = \int_{l_{m+1:t}} p(Y_t=y|H_m,L_{m+1:t}=l_{m+1:t}, A_{m:t}=g(H_{m:t})) \nonumber \\
& \quad\quad\quad\quad \times \prod_{j=m+1}^t p(L_j = l_j|H_m,L_{m+1:j-1}=l_{m+1:j-1},\nonumber\\
& \quad\quad\quad\quad\quad\quad\quad\quad A_{m,j-1}=g(H_m,l_{m+1:j-1})).
\end{align}
}%


We can approximate this integral through Monte-Carlo simulation. 

G-computation algorithm requires the ability to simulate from joint conditional distributions $p(L_{t} |\bar{L}_{t-1},\bar{A}_{t-1})$ of the covariates given patient history.
These conditional distributions need to be estimated from data. Most implementations of g-computation use generalized linear regression models (GLMs) to estimate the conditional distributions of the covariates. These models often do not capture the complex temporal dependencies in the patient data. We propose the G-Transformer for this task.

\subsection{G-Transformer Architecture}
\begin{figure}[thb]
    \centering
    \includegraphics[scale=0.34]{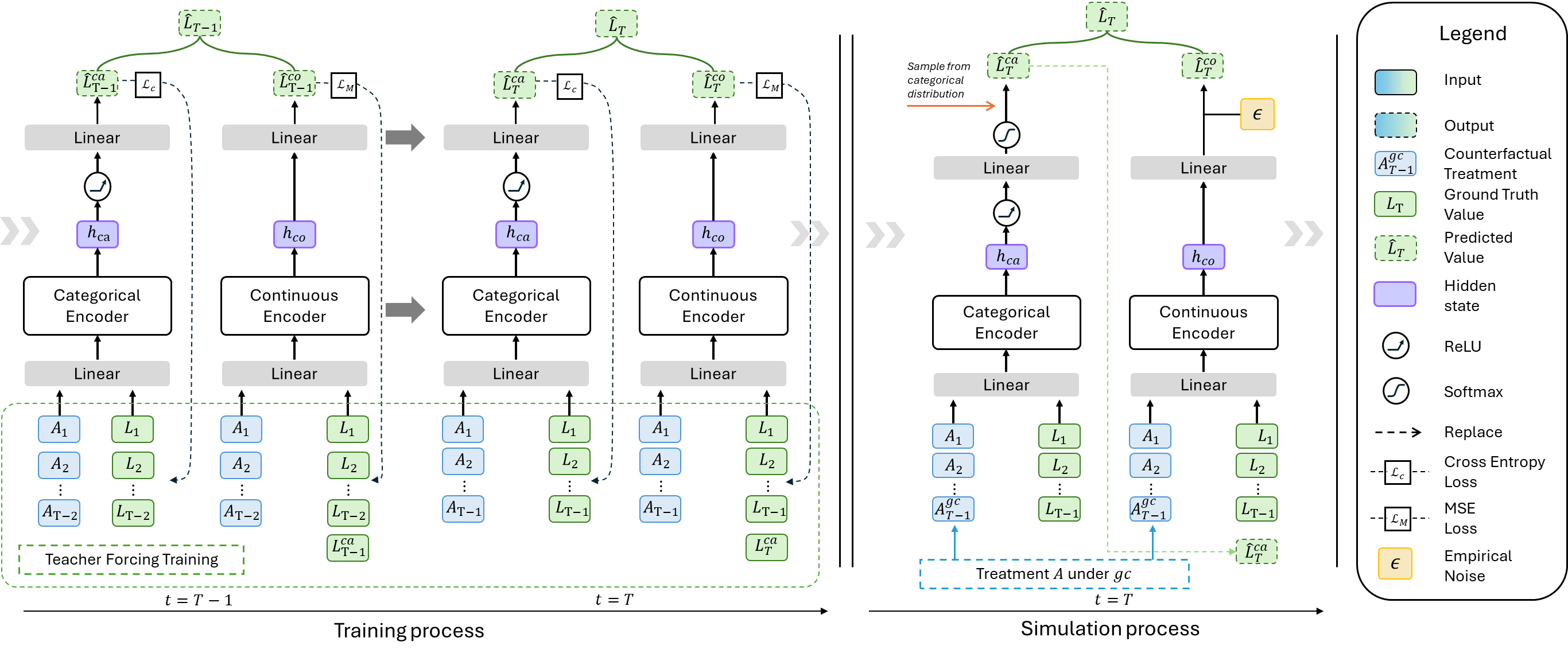}
    \caption{\small{G-Transformer Architecture. In this diagram, G-Transformer uses two distinct encoders for categorical ($L^{ca}$) and continuous variables ($L^{co}$) respectively. The diagram illustrates (i) the training process through time steps T-1 and T, and (ii) the counterfactual simulation process starting at time T, with the counterfactual strategy $gc$ initiated at the end of time T-1.  Teacher forcing is used during training. } }
    \label{fig:model structure}
\end{figure}
We utilized two Transformer encoders as the sequential model to separately learn hidden representations for continuous and categorical covariates in G-Transformer. We use the teacher forcing in the training process.  
Although we have presented our model architecture with two Transformer encoders—one for continuous variables and another for categorical variables— G-Transformer can be extended to support multiple encoders. For instance, a separate Transformer encoder can be used to model the conditional distribution of each individual covariate.

Let $L_t^0,...,L_t^p$ denote $p$ components of the vector of covariates $L_t$. When we calculate joint conditional distributions $p(L_t|\bar L_{t-1},\bar A_{t-1})$ at time t, we leverage the conditional probability identity as in \citep{GNet2021}:


\begin{align}
p(L_t | \bar L_{t-1}, \bar A_{t-1}) &= p(L^0_t | \bar L_{t-1}, \bar A_{t-1}) \notag \\ 
&\times p(L^1_t | L^0_t, \bar L_{t-1}, \bar A_{t-1})  \notag \\ 
&\times \cdots \notag \\ 
&\times p(L^{P-1}_t | L^{0}_t, \ldots, L^{P-2}_t, \bar L_{t-1}, \bar A_{t-1})
\end{align}

In this work, unless otherwise noted, we divide covariates into two groups, 
one for the categorical variables $L_t^{ca}$, and the other for the continuous variables $L_t^{co}$. For simplicity,  our description assumes categorical variables are simulated before continuous variables within each time step. However, our implementation does not impose any  specific ordering requirements among the covariates.





We use two Transformer encoder models to perform representation learning for categorical variables and continuous variables respectively. At each time step $t$, we can compute representation of patient history $R_t=r_t(\bar L_t,\bar A_t;\theta)$, where $\theta$ denotes the learnable parameters from model $r_t$. We will introduce the specific content of model $r_t$ in the following section.  In this work, the outcome $Y_t$ can be deemed as a covariate in the vector $L_{t+1}$. For example, the outcome of interest can be blood pressure in the next time step.  

\textbf{Representation Learning.} For the input of categorical encoder, we concatenate all the historical treatments $\bar A_t$ and covariates $\bar L_t$ together, and use a linear layer to obtain their combined representation $h_{ca}^r$:
\begin{equation}
    h_{ca}^r=Linear(concat(\bar A_t,\bar L_t))
\end{equation}
Afterwards, we feed the combined representations of treatments and outcomes $h_{ca}^r$ into a Transformer encoder to obtain the final representation of past covariate history:
\begin{equation}
    h_{ca} = Transformer\: Encoder_{ca}(h_{ca}^r)
\end{equation}
The internal architecture of the Transformer encoder sublayers follows the original Transformer paper \citep{vaswani2017attention}, including a multi-head self-attention layer, feed-forward network, and layer normalization mechanism. After obtaining $h_{ca}$, we map the hidden representation to the predicted future time step $\hat L^{ca}_{t+1}$ through a linear layer, followed by a softmax layer for getting probability $p^{ca}$ of categorical variables before loss calculation.
\begin{equation}
    \hat {L}_{t+1}^{ca}=ReLU(Linear(h_{ca}))
\end{equation}

During the training phase, we use \textbf{teacher forcing} to guide the training of the categorical and continuous encoders. In addition to the patient's historical sequences $(\bar A_t,\bar L_t)$, we also use observed value $L_{t+1}^{ca}$ as input into the continuous encoder:
\begin{equation}
    h_{co}^r=Linear(concat(\bar A_t,\bar L_t, L_{t+1}^{ca})).
\end{equation}
As with categorical variables, we use a Transformer encoder to obtain a hidden representation $h_{co}$ that incorporates the preceding variables. Then we utilize a linear layer to modify the hidden state to generate the final output $ \hat {L}_{t+1}^{co}$:
\begin{align}
    h_{co} = T&ransformer\: Encoder_{co}(h_{co}^r) \\
    &\hat {L}_{t+1}^{co} = Linear(h_{co})
\end{align}
    
\subsection{Training of G-Transformer}

Since g-computation models the joint conditional probability distribution  of the next time step $p(L_t|\bar L_{t-1},\bar A_{t-1})$ given the observable sequence, it aligns more closely with the modeling methods used in RNN-like regression models. However, due to the Transformer's nature as a seq-2-seq model, with its multi-head self-attention mechanism, it is typically used to generate an entire sequence at a time.  We developed a customized training procedure that is aligned with the g-computation algorithm (see Algorithm \ref{algorithm:training}). 
As a comparison, we present results from alternative  training methods on our counterfactual prediction task,  including the conventional Seq-2-Seq approach, and the results are shown in the Appendix \ref{app:1a}.

We use cross-entropy loss and MSE (Mean Squared Error) loss to optimize the prediction results for categorical and continuous variables, respectively. For categorical variables:
\begin{equation}
\label{eq:ce}
    \mathcal{L}_{ce} = -\frac{1}{N (K-m) D_{ca} }\sum_{i=1}^{N} \sum_{d=1}^{D_{ca}} \sum_{t=m}^{K} \sum_{c=1}^{C}  L^{c,ca}_{idt} log( p_{idt}^{c,ca})
\end{equation}
where $D_{ca}$ denotes the number of categorical variables, $N$ denotes the number of paitents, $\theta$ denotes the model parameters, and $C$ denotes the number of classes in this categorical variable. 
We assume the class label of $L^{ca}_{d}$ is represented as a one-hot encoded vector 
$(L^{1}_{d}, L^{2}_{d},..., L^{C}_{d})$ where $L^c_{d} = 1$ for the correct class and 0 for all other classes.  We separately calculated the cross-entropy losses of different variables and computed their averages to account for the needs of multi-class variables. For continuous variables, we utilize the MSE loss:
\begin{equation}
\label{eq:mse}
    \mathcal{L}_{mse} =\frac{1}{N(K-m)D_{co}} \sum_{i=1}^{N}\sum_{d=1}^{D_{co}} \sum_{t=m}^{K}||  L_{t,i}^{d,co} - \hat L_{t,i}^{d,co} ||^2
\end{equation}
where $\hat L_{t,i}^{co}$ denotes the output of continuous encoder part of G-Transformer for patient $i$ in time $t$, $N$ denotes number of patients, $K$ denotes length of time and $D_{co}$ denotes the number of continuous variables.  Algorithm \ref{algorithm:training} illustrates the overall training process of G-Transformer.

\begin{algorithm2e}
\caption{G-Transformer training algorithm}
\label{algorithm:training}
\DontPrintSemicolon
\KwIn{Training set \( \mathcal{D} \) data tuples: \( \mathcal{D} = \{ (\bar{L},\bar{A}) \} \) ; The number of the prediction length \( K \); The number of the observed historical length \( m \); Model parameters \( \theta_{ca} \) and \(\theta_{co}\); The number of the learning epochs \(E\),.}
\While{\(e<E\)}{
\For{\( t = m \) \KwTo \(  K \)}{
    Use categorical encoder to compute the \(\hat{L}_{t}^{ca} =f_{ca}(\bar{L}_{1:t-1}, \bar{A}_{1:t-1};\theta_{ca})\). \;
    Use continuous encoder to compute the \(\hat{L}_{t}^{co} =f_{co}(\{\bar{L}_{1:t-1}, \bar{A}_{1:t-1},  {L}_{t}^{ca}\};\theta_{co})\). \;
    Save the \(\hat{L}_{t}^{co},\hat{L}_{t}^{ca}\) for loss calculation.\;
}
Concatenate the results of all predicted time steps \(\hat{L}_{m:K}^{ca}=\{\hat{L}_{m}^{ca},...,\hat{L}_{K}^{ca}\}\) and \(\hat{L}_{m:K}^{co}=\{\hat{L}_{m}^{co},...,\hat{L}_{K}^{co}\}\). \; 
Update model parameters \(\theta_{ca}\) and \(\theta_{co}\) by minimizing (\ref{eq:ce}) and (\ref{eq:mse}) with gradient descent. \; 
}

\KwRet{ updated model parameters \(\theta_{ca}\) and \(\theta_{co}\)}
\end{algorithm2e}

\subsection{Simulation}


After having  trained a sequential model $f(R;\theta)$, we can perform Monte-Carlo simulations, and sequentially simulate from $p(L_t^p|\bar{L}_{t-1},L_t^0,\ldots,L_t^{p-1},\bar{A}_{t-1})$  as follows. If $L_t^p$ is multinomial, its conditional expectation defines its conditional density. If $L_t^p$ has a continuous density, we simulate from its conditional distribution as follows.  Without making parametric assumptions, we simulate from:
\begin{equation}
L_t^{p} \simeq   \mathbb{ \hat E}[(L_t^{p} | L_t^0, ..., L_t^{p-1},\bar L_{t-1},\bar A_{t-1})]+\epsilon_t^p
\end{equation}
where $\epsilon_t^p$ is a draw from the empirical distribution of the residuals $L^p_t-\hat L^p_t$ in a holdout set. 
We use Algorithm \ref{alg:gcomp} for each time $t = \{m, ...., K\}$ to perform multi-timestep simulations. 
Figure \ref{fig:model structure} illustrates the simulation process. Different from the training process, since we do not have access to the ground truth, we sample from the learned conditional distribution when performing counterfactual predictions.  For categorical variables,  the predicted probabilities from $\hat L^{ca}$, produced by the categorical encoder, are used to parameterize a categorical distribution (or a Bernoulli for binary indicator variables). We then sample from the resulting distribution. For the continuous variables, we added noise sampled from the empirical distribution collected from the validation set to the output of the continuous encoder.  

\begin{algorithm2e}[h!]
\DontPrintSemicolon
\SetKwInOut{Input}{Input}
\SetKwInOut{Output}{Output}
\caption{Simulation (one time-step)}
\label{alg:gcomp}
Set $a_m^*=g_m(H_m)$\; 
Sample $l_{m+1}^*$ from $p(L_{m+1} |H_m,A_{m}=a_{m}^*)$\; 
Set $a_{m+1}^*=g_m(H_m,l_{m+1}^*,a_{m}^*)$\; 
Sample $l_{m+2}^*$ from $p(L_{m+2}|H_{m},L_{m+1}=l_{m+1}^*,A_{m}=a_{m}^*,A_{m+1}=a_{m+1}^*)$\;
Continue simulations through time $K$
\end{algorithm2e}

We repeat the g-computation simulation (Algorithm \ref{alg:gcomp})
$M$ times, where $M$ is the number of Monte-Carlo simulations. At the end of this process, we have $M$ simulated draws of the counterfactual outcome for each time $t=\{m,\ldots,K\}$, where $m$ is beginning time step of simulation and $K$ is the end time step of simulation. For each $t$, the empirical distribution of these draws constitutes a Monte-Carlo approximation of the counterfactual outcome distribution (\ref{target2}). The sample averages of the draws at each time $t$ are an estimate of the conditional expectations (\ref{target}) and can serve as point predictions for $Y_t(\bar{A}_{m-1},\underline{g}_m)$ in a patient with history $H_m$.




\subsection{Evaluation}

To evaluate G-Transformer, we use simulated data in which counterfactual ground truth can be known. Specifically, we use CVSim \citep{CVSim2010}, a well-established mechanistic model of the cardiovascular system, to simulate counterfactual patient trajectories under various dynamic fluid and vasopressor administration strategies. Additionally, using a simulated tumor growth data set, we  compare G-Transformer with recently introduced G-Net \citep{GNet2021}, Recurrent Neural Networks (CRN) \citep{bica2020crn}, a Recurrent Marginal Structural Network (R-MSN) \citep{Lim2018}, a recurrent neural network implementation of a history adjusted marginal structural model for estimating static time-varying treatment effects, and Causal Transformer \citep{melnychuk2022causal}.  

We applied our proposed G-Transformer approach to predicting outcomes of sepsis patients in the ICU under alternative fluid resuscitation treatment regimes (e.g. aggressive vs. conservative) using de-identified real-world intensive care units (ICU) data from the MIMIC-IV \citep{mimic-iv} publicly available from PhysioNet \cite{GoldbergerPhysioNet2000}.  
We quantitatively evaluate G-Transformer's performance in predicting patient trajectories under observational treatment regimes learned from the ICU data. This entails using the G-Transformer architecture to predict treatment at each time-step under the observational treatment regime in the observational data. Under predictive check, the treatments are from G-Transformer predicted actions conditioned on patient history, and we evaluate by comparing the predicted time-varying outcomes (averaged across $M$ Monte-Carlo simulations) with the actual observed trajectories from individual patients.

\section{Experiments on Simulated Data from Mechanistic Models}

\subsection{Cancer Growth Experiments and Results}

\paragraph{Cancer Growth Data Generation} As in \cite{Lim2018,bica2020crn,GNet2021,melnychuk2022causal}, we generate simulated `observational' data from a pharmacokinetic-pharmacodynamic (PK/PD) model of tumor growth under a stochastic regime \citep{cancer_sim}.  In this simulation, chemotherapy and radiation therapy comprise a two dimensional time-varying treatment impacting tumor growth. Under the observational regime, probability of receiving each treatment at each time depends on volume history, so there is time-varying confounding. 
 

 
\begin{table}[!hbt!]
 \caption{\small{ Cancer growth data: Percent RMSE for various prediction horizons. A = Overall. CT = Causal Transformer. Best performing models in bold.}} 
\label{cancer_growth_simulation}
    \centering
    \begin{tabular}{c|c|ccccccc} 
    \toprule
        {} & {} & rMSN & CRN & Linear & G-Net & CT & G-Transformer  \\
        {} &  $t$  &\tiny{--}  &\tiny{--}  & \small{g-comp} & \tiny{--} &\tiny{--}  &  \\
        \midrule

        No  & \small{1} & 1.13 & 1.00 & 0.63  & \textbf{0.25} & 0.67  &  0.26 \\
        {Treat} & \small{2} & 1.24 & 1.20 & 1.21  & 0.47 & 0.65  & \textbf{0.44} \\
        {} & \small{3} & 1.85 & 1.49 & 1.78  & 0.72 & 0.63  & \textbf{0.55} \\
        {} & \small{4} & 2.60 & 1.78 & 2.35  & 1.01 & \textbf{0.63}  & 0.66\\
        {} & \small{A} & 1.68 & 1.40 & 1.62  & 0.67 & 0.65  & \textbf{0.48}\\ \hline

        Radio  & \small{1} & 5.27 & 4.91 & 7.14  & 3.29 & 5.15  & \textbf{2.74}\\
        {} & \small{2} & 5.38 & 4.92 & 7.43  & 3.14 & 2.98  & \textbf{2.50}\\
        {} & \small{3} & 5.13 & 4.94 & 7.05  & 3.02 & 2.72  & \textbf{2.40}\\
        {} & \small{4} & 4.81 & 4.92 & 6.50  & 3.00 & 2.53  & \textbf{2.52}\\
        {} & \small{A} & 5.15 & 4.92 & 7.04  & 3.11 & 3.35  & \textbf{2.54}\\ \hline

        Chemo  & \small{1} & 1.42 & 1.04 & 1.58  & \textbf{0.34} & 0.66& 0.36\\
        {} & \small{2} & 1.27 & 1.09 & 3.14  & 0.63 & \textbf{0.50}  & 1.16\\
        {} & \small{3} & 1.46 & 1.03 & 4.52  & 0.84 & \textbf{0.44}  & 1.95\\
        {} & \small{4} & 1.69 & 1.02 & 5.47  & 0.89 & \textbf{0.41}  & 2.45\\
        {} & \small{A} & 1.47 & 1.05 & 3.96  & 0.71 & \textbf{0.50}  &  1.48\\\hline

        Radio  & \small{1} & 4.76 & 4.66 & 7.76  & 3.10 & 8.40  & \textbf{2.48}\\
        {Chemo} & \small{2} & 3.59 & 4.36 & 7.32  & 2.28 & 7.72  & \textbf{2.04}\\
        {} & \small{3} & 2.76 & 3.65 & 6.13  & \textbf{1.50} & 7.79  & 1.54\\
        {} & \small{4} & 2.30 & 2.95 & 4.88  & 1.18 & 7.67  & \textbf{1.10}\\
        {} & \small{A} & 3.48 & 3.96 & 6.62  & 2.15 & 7.90  & \textbf{1.79}\\\hline
    \end{tabular}
\end{table}


 We use the same experiment settings and data generation procedure as reported in \cite{GNet2021}. Briefly, we generated four test sets in which four counterfactual regimes  were followed for the final four time points in the test set: (1) give only radiotherapy, (2) only chemotherapy, (3) both chemotherapy and radiotherapy, and (4) no treatment.  We generated 10,000 trajectories for training, and 1,000 simulated trajectories for hyperparameter optimization (validation data), with another 1,000 for  testing. See \cite{GNet2021} for full details of the data generation process.  These counterfactual regimes are static time-varying regimes (in which treatments do not depend on prior patient history), as rMSN, CRN, and Causal Transformer are intended to estimate counterfactuals under static time-varying regimes.

Table \ref{cancer_growth_simulation} presents the percent root mean square error (RMSE) of predictions from rMSN, CRN, linear implementation of g-computation, G-Net, Causal Transformer, and G-Transformer in the final four time steps when counterfactual strategies were in effect, conditioned upon previous time points. The raw RMSE values were divided by the maximum possible tumor volume, 1150 $cm^2$, as in \cite{Lim2018}, to calculate the percent RMSE. Details of the model training and hyper-parameter settings are described in the Appendix.

We observe that G-Transformer achieved the best overall RMSE performance in three of the four cancer growth datasets, demonstrating its advantages in counterfactual prediction task compared to other state-of-the-art deep learning models, including rMSN, CRN, G-Net and Causal Transformer.  Causal Transformer outperformed other models under the chemotherapy (Chemo) counterfactual regime, but fall short in other counterfactual regimes.


\subsection{CVSim Experiments and Results}

\paragraph{CVSim Data Generation}  To evaluate counterfactual predictions, it is necessary to use simulated data in which counterfactual ground truth for outcomes under alternative treatment strategies is known. To this end, we performed experiments on data generated by CVSim, a program that simulates the dynamics of the human cardiovascular system \citep{CVSim2010}. We used a CVSim 6-compartment circulatory model which takes as input 28 variables that together govern a hemodynamic system. We built on CVSim  by adding stochastic components and interventions for the purposes of evaluating our counterfactual simulators. 

We generated an `observational' dataset $D_{o}$ under treatment regime $g_{o}$ and two `counterfactual' datasets $D_{c1}$ and $D_{c2}$ under treatment regimes $g_{c1}$ and $g_{c2}$. The data generating processes producing $D_{o}$ and $D_{cj}$ were the same except for the treatment assignment rules. For each $j$, $g_{cj}$ was identical to $g_{o}$ for the first $m-1$ simulation time steps before diverging to a different treatment rule for time steps $m$ to $K$. Full details are in the Appendix \ref{appendix:cvsim_data}.

\paragraph{Experiment} 



As in \cite{GNet2021}, we generated a total of 12,000 trajectories in  $D_{o}$ ($N_{o}=12,000$), of which 80\% were used for training, and the remaining 20\% for validation.  For testing, we generated 1000 observations in the $D_{cj}$ datasets ($N_{c}=1000$).  We included a total of 20 output variables (i.e. two treatment variables and 18 covariates influencing treatment assignment under $g_o$) from CVSim to construct $D_{o}$ and $D_{cj}$; each trajectory is of length 66 time steps (d=20, K=66). In  each $D_{cj}$, the switching time point $m$ from $g_{o}$ to $g_{c}$ is fixed at 34 for all trajectories ($m=34$).  

Given observed covariate history through 34 time steps and treatment history through 33 time steps of each trajectory in each $D_{cj}$, we computed both the population-level and individual-level RMSE, as well as the calibration, of counterfactual predictions from G-Transformer and its counterparts for time steps 35 to 66. Moreover, we presented the estimated and actual population-level average trajectories under \textit{g$_{c1}$} and \textit{g$_{c2}$} for selected variables. 



\paragraph{CVSim Counterfactual Prediction Performance in RMSE.} We use RMSE to assess our model's accuracy in counterfactual prediction for continuous variables in CVSim data.  Individual-level RMSE is calculated by first determining the square root of the mean squared differences between each patient's observed and predicted trajectories. We also report population-level RMSE, by comparing the average population-level trajectory  of the predicted and the ground-truth.


For G-Transformer, G-Net and the linear implementation of g-computation, the predicted trajectory  is averaged from 100 Monte Carlo simulations. 
The RMSE values across all patients are then averaged to derive individual-level RMSE.

\begin{table*}[th!]
\centering 
\caption{\small{CVSim: Counterfactual prediction under \textit{g$_{c1}$} and \textit{g$_{c2}$}. Performance of various models in predicting remaining 32-hour patient covariate trajectories based on first 34-hour patient covariate trajectories. Values reported represent the population-level and individual-level RMSE between the predicted and counterfactual trajectories. * Causal Transformer modified to update treatment assignment at each time-step under gc${_1}$ and gc$_{2}$.}}
\label{tab:cvsim_pop_and_individual_RMSE_gc1_and_gc2}
\small
\begin{tabular}{ c|p{2.3cm} p{2.3cm}|p{2.3cm} p{2.3cm} } 
 \hline 
 \multicolumn{1}{c|}{} & \multicolumn{2}{c|}{Population Level} & \multicolumn{2}{c}{Individual Level} \\
 \textbf{Model} & \textbf{RMSE$_{gc1}$} & \textbf{RMSE$_{gc2}$}  & \textbf{RMSE$_{gc1}$} & \textbf{RMSE$_{gc2}$}\\
 \hline \hline
  \textbf{Causal Transformer$^{*}$} & 0.576 $\pm$ 0.078  & 1.323 $\pm$ 0.257 & 1.473 $\pm$ 0.272 & 1.694 $\pm$ 0.407 \\
  \textbf{{Linear}(g-comp)} & 0.255 $\pm$ 0.001&  0.442 $\pm$ 0.002 & 1.103 $\pm$ 0.001& 1.307 $\pm$ 0.001 \\
 \textbf{G-Net} & 0.066 $\pm$ 0.001 & 0.329 $\pm$ 0.101 & 1.020 $\pm$ 0.002 & 1.220 $\pm$ 0.048 \\
 \textbf{G-Transformer} & \textbf{0.055 $\pm$ 0.006}  & \textbf{0.299 $\pm$0.110} &  \textbf{1.019 $\pm$ 0.002} & \textbf{1.210 $\pm$ 0.045} \\
 \hline
\end{tabular}
\end{table*}

Table \ref{tab:cvsim_pop_and_individual_RMSE_gc1_and_gc2} presents a comparison of individual-level and population-level RMSE among linear implementation of g-computation, G-Net, Causal Transformer, and G-Transformer under \textit{g$_{c1}$} and \textit{g$_{c2}$}.
The G-Transformer outperforms linear implementation of g-computation and G-Net, and considerably outperforms Causal Transformer in this counterfactual prediction task.     Table \ref{cvsim_individual_level_rmse_timestamp} in the Appendix \ref{appendix:cvsim_ts_rmse} presents performance in RMSE over-time. It should be noted that Causal Transformer \cite{melnychuk2022causal} can estimate counterfactual outcomes under time-varying treatments, but it is not designed to estimate counterfactual outcomes under \textit{dynamic} treatment regimes, where treatments depend on past covariate history during counterfactual prediction. Nevertheless, we modified the Causal Transformer's implementation to update the sequential treatment assignment at each time-step according to the specified dynamic treatment rule depending on the past covariates, and present its performance here as a baseline.




\paragraph{Population-Level Trajectories}

\begin{figure} [h!]
  \begin{subfigure}{0.32\linewidth}
    \centering
    \includegraphics[width=\linewidth]{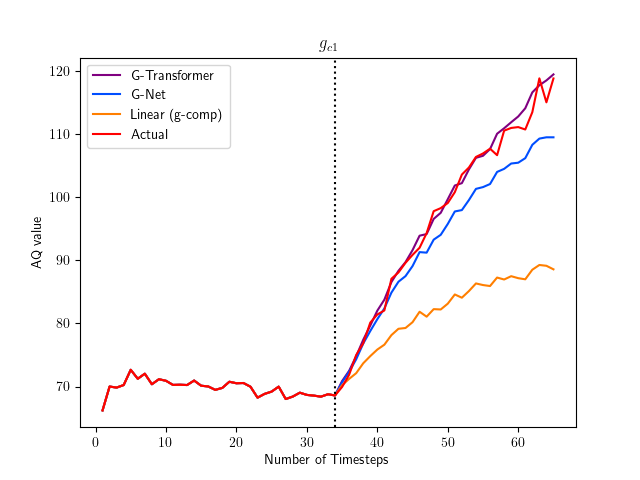}
    \caption{AQ ($gc_{1}$)}
    \label{fig:cvsim_gc1_aq}
  \end{subfigure}
 \begin{subfigure}{0.32\linewidth}
    \centering
    \includegraphics[width=\linewidth]{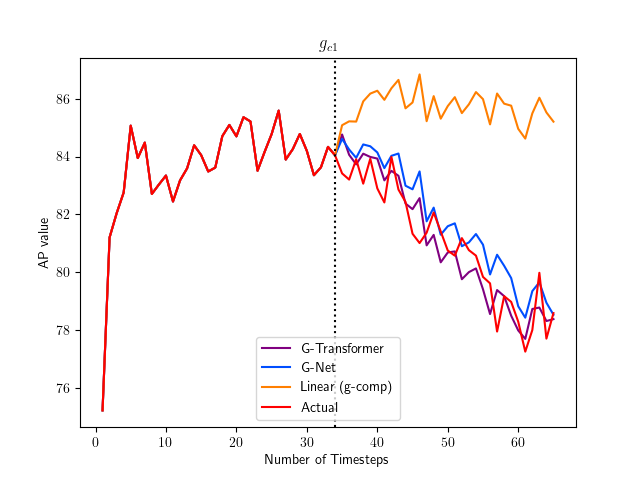}
    \caption{AP ($gc_{1}$)}
    \label{fig:cvsim_gc1_ap}
  \end{subfigure}
  \begin{subfigure}{0.32\linewidth}
    \centering
    \includegraphics[width=\linewidth]{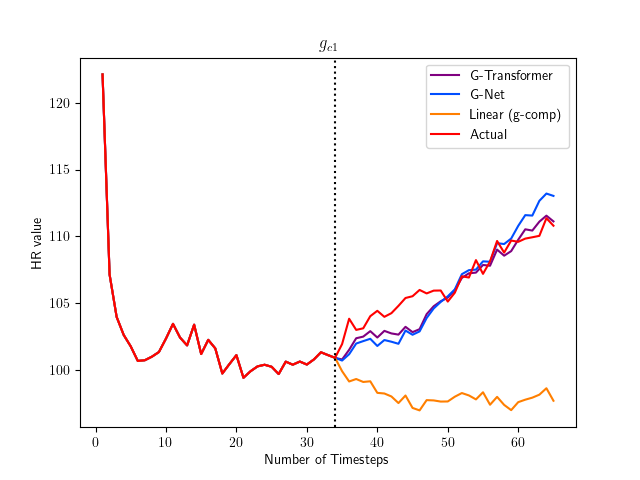}
    \caption{HR ($gc_{1}$)}
    \label{fig:cvsim_gc1_hr}
  \end{subfigure} \\
    \begin{subfigure}{0.32\linewidth}
    \centering
    \includegraphics[width=\linewidth]{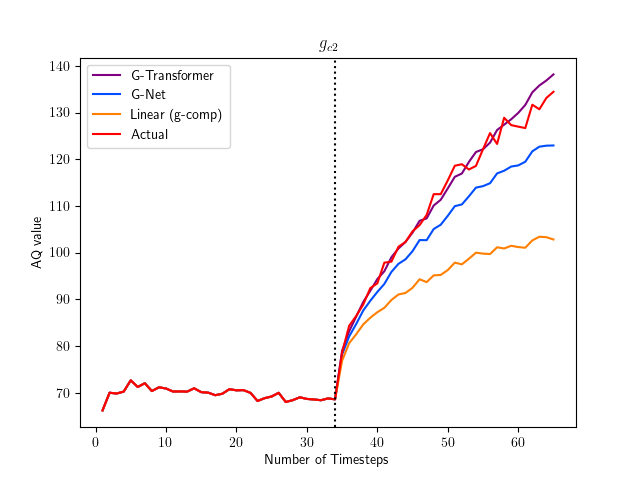}
    \caption{AQ ($gc_{2}$)}
    \label{fig:cvsim_gc2_aq}
  \end{subfigure}
 \begin{subfigure}{0.32\linewidth}
    \centering
    \includegraphics[width=\linewidth]{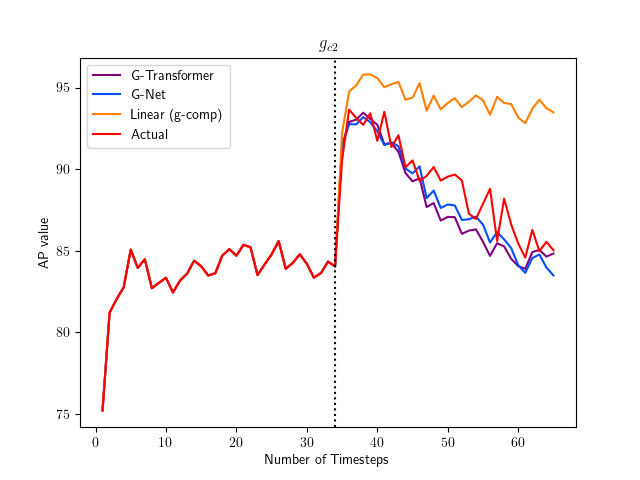}
    \caption{AP ($gc_{2}$)}
    \label{fig:cvsim_gc2_ap}
  \end{subfigure}
  \begin{subfigure}{0.32\linewidth}
    \centering
    \includegraphics[width=\linewidth]{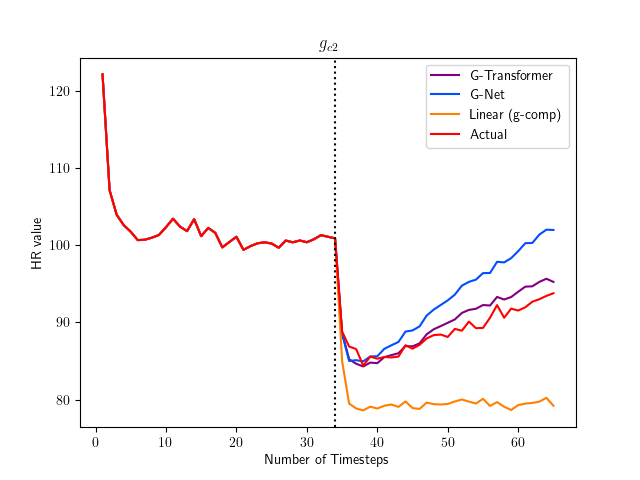}
    \caption{HR ($gc_{2}$)}
    \label{fig:cvsim_gc2_hr}
  \end{subfigure} \\
  \caption{\small{Plots \subref{fig:cvsim_gc1_aq} to \subref{fig:cvsim_gc2_hr} show the estimated and actual population average trajectories for three selected variables: arterial flow (AQ), arterial pressure (AP), and heart rate (HR), according to three models' predictions under \textit{g$_{c1}$} vs \textit{g$_{c2}$}. 
  }}
\label{fig:cvsim_AQ_pop_plot}
\end{figure}

The G-Transformer, G-Net, and linear implementation of g-computation can be used to quantitatively demonstrate counterfactual outcomes through population average trajectories. The estimated and actual population-level average trajectories under \textit{g$_{c1}$} and \textit{g$_{c2}$} for selected variables, specifically arterial flow (AQ), arterial pressure (AP), and heart rate (HR), are presented in Figure \ref{fig:cvsim_AQ_pop_plot} \subref{fig:cvsim_gc1_aq} to \subref{fig:cvsim_gc2_hr}.
The plots show that G-Transformer outperforms G-Net and linear implementation of g-computation, and more accurately predict AQ, AP, and HR counterfactual trajectories from the population level.



\paragraph{Calibration}



\begin{figure}
 \centering
  \begin{subfigure}{0.45\linewidth}
    \centering
    \includegraphics[width=\linewidth]{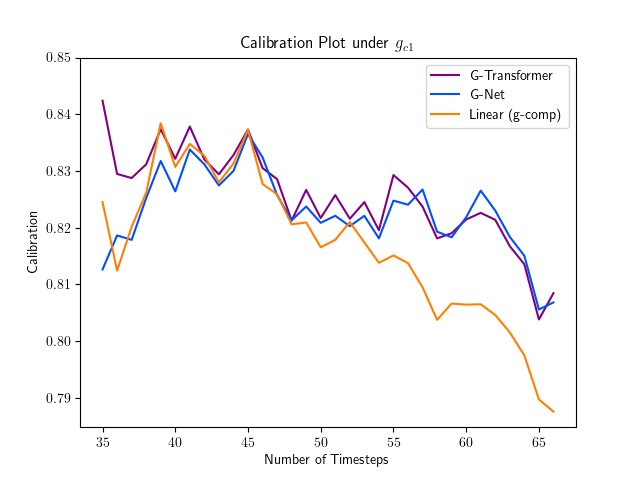}
    \caption{Calibration under $gc_{1}$}
    \label{fig:subfiga}
  \end{subfigure}
  \begin{subfigure}{0.45\linewidth}
    \centering
    \includegraphics[width=\linewidth]{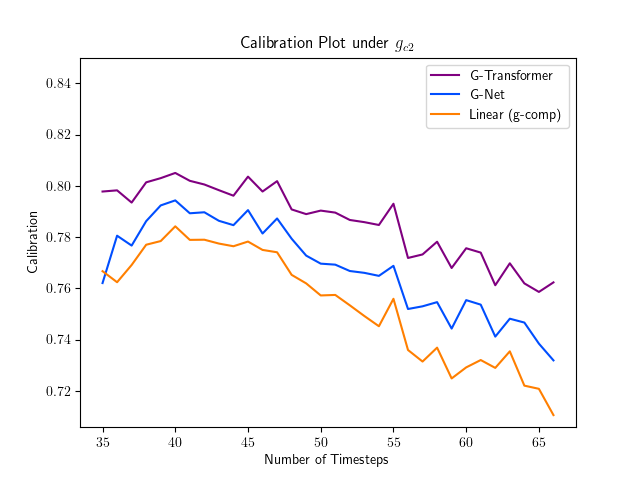}
    \caption{Calibration under $gc_{2}$}
    \label{fig:subfigb}
  \end{subfigure}
  \caption{\small{Calibration over time for $g_{c1}$ and $g_{c2}$}}
  \label{fig:cvsim_calibration}
\end{figure}

We assess the calibration of a G-Transformer, G-Net, or linear implementation of g-computation, denoted as $G$, as follows. Given lower and upper quantiles $\alpha_{low}$ and $\alpha_{high}$, the calibration measures the frequency with which the actual counterfactual covariate $L_{ti}^{h,cj}$ is between the $\alpha_{low}$ and $\alpha_{high}$ quantiles of the $M$ simulations $\{\tilde{L}_{ti}^{h,cj}(H_{mi}^{cj},G,k):k\in1:M\}$. If this frequency is approximately $\alpha_{high}-\alpha_{low}$, then $G$ is well calibrated.

Figure \ref{fig:cvsim_calibration} depicts calibration for G-Transformer, G-Net, and linear implementation of g-computation with lower and upper quantiles of 0.05 and 0.95. G-Transformer generally performed better than G-Net and linear implementation of g-computation over all time steps.

\section{Experiments on MIMIC Data}

\subsection{MIMIC Cohort} 

The data employed in this study was extracted from the Medical Information Mart for Intensive Care IV database (MIMIC-IV v1.0), containing medical records from more than 523,500 hospital admissions and 76,500 ICU stays at the Beth Israel Deaconess Medical Center (BIDMC) between 2008 and 2019 \cite{mimic}. Our cohort consists of ICU patient  identified as septic under the Third International Consensus Definitions for Sepsis and Septic Shock (Sepsis-3) \cite{sepsis3-def} and did not meet exclusion criteria (see Appendix \ref{appendix:mimic_data} for details).
Our final sepsis cohort consisted of 8,934 total patients which we further split into 7,147 patients, 893 patients, and 894 patients for the training, validation, and test data respectively.

\subsection{Study Design} 

Since there is no ground-truth observations available for the MIMIC data for counterfactual predictions, we first assess our models' performance through a ``predictive check'' using data collected during the first 24 hours of patients' stay in the ICU. This involved conducting Monte Carlo simulations and projecting forward patient covariate trajectories within the test set under the observational regime. Subsequently, we compared these simulated trajectories (averaged across 100 Monte Carlo simulations per patient) to the ground-truth data.  Following predictive check, we assess our model in performing counterfactual predictions. In the MIMIC experiment, we use a separate Transformer encoder for each covariate in G-Transformer. For G-Net, we use a separate LSTM for each covariate. 

\subsection{Predictive Checks} 

For each patient in the test set, we use G-Transformer and other baseline approaches to simulate their trajectories under the observational treatment regime from hour k up to hour 24, conditioned on the observed covariates and treatment history up to and including hour k-1.  Under predictive check in G-Transformer, we use a Transformer encoder to predict the treatment action at each time step under the observational treatment regime.  Similarly, for G-Net, we use an LSTM model to predict the treatment at each time step during predictive checks. We compared the trajectories between predicted and ground-truth under G-Transformer, G-Net, and Causal Transformer. We use individual-level RMSE as a metric.
Since patient trajectories can be censored due to events such as death and release in MIMIC data,
individual-level RMSE is adjusted to account for error due to over- and under-prediction in the trajectory length as in 
\cite{sucounterfactual} (see Appendix \ref{appendix:mimic_corrections} for details).


We also compared the models' performance in predicting the occurrence of adverse clinical outcomes during the first 24-hours patients were in the ICU. 
To generate a probability for the 24-hour window, we first assigned a binary label for each Monte Carlo simulation based on whether the outcome is predicted to have occurred: $1$ if outcome of interest was predicted to occur between timestamps $k$ to $24$, and $0$ otherwise. Then the average across the Monte Carlo simulations per patient was used as the predicted probability for that patient. The ground truth was determined in the same manner by looking for the presence of the outcome of interest in the ground-truth trajectory. 


\subsection{Predictive Check: Individual-Level RMSE} \label{mimic_predictive_check}



Table \ref{tab:mimic_2_box_RMSE} presents the predictive check on individual-level RMSE for continuous variables. We aim to assess the performance of G-Transformer, G-Net, and Causal Transformer in predicting 24-hour patient covariate trajectories, starting at time $k$ conditioned on covariates from the previous $k-1$ hours in the ICU. 
Values reported represent the RMSE of continuous covariates between the predicted and ground-truth trajectories.  To account for mismatch in trajectory length due to over- and under-prediction of death and release time, for both predicted and actual trajectories, depending on nature of covariates, post-death timesteps are padded with one of the following values: normalized zero, normalized minimum value in dataset, or normalized logarithmic zero. The post-release timesteps are consistently padded with normalized population mean.

Results in Table \ref{tab:mimic_2_box_RMSE} indicate that
G-Transformer consistently outperforms G-Net and Causal Transformer across different values of $k$. 
 Although Causal Transformer \cite{melnychuk2022causal} is not intended to estimate counterfactual outcomes under dynamic treatment regimes, nevertheless, we present the performance from a modified version of Causal Transformer (where treatments depend on past covariate history during simulations) here as a baseline.
\begin{table*}[htb!]
\centering 
\caption{\small{MIMIC-IV Experiments: Predictive check on continuous variables at individual-level RMSE. $k$ is the simulation start time relative to the ICU admission time. Performance in predicting 24-hour patient covariate trajectories, starting at hour $k$ conditioned on covariates from the previous $k-1$ timesteps of post-ICU admission. Values reported represent the individual-level RMSE of continuous covariates between the predicted and ground-truth trajectories. Individual-level RMSE is adjusted to account for error due to over- and under-prediction in the trajectory length as a result of death and discharge. *Causal Transformer modified to update treatment assignment at each time-step based on predicted action.} }

\label{tab:mimic_2_box_RMSE}
\small
\begin{tabular}{c|cc} 
 \hline 
 \multicolumn{1}{c|}{} & \multicolumn{2}{c}{Individual Level RMSE} \\
 \cline{2-3}
 \multicolumn{1}{c|}{\textbf{Model / k (hours)}} &  \textbf{2} & \textbf{6} \\
 \hline \hline
 \textbf{Causal Transformer$^{*}$} & 14.37 & 15.73 \\
 \textbf{G-Net}  & 7.017 & 7.670\\
 \textbf{G-Transformer} & \textbf{6.993} & \textbf{7.643} \\
 \hline
\end{tabular}
\end{table*}


\paragraph{Predictive Check: an Illustrative Example from G-Transformer's Monte Carlo Simulations} 

\begin{figure}
 \centering
  \begin{subfigure}{0.4\linewidth}
    \centering
    \includegraphics[width=\linewidth]    {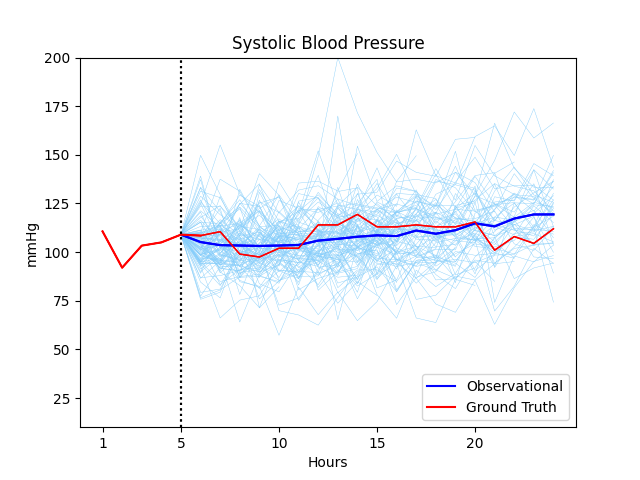}
    \caption{SBP}
    \label{fig:subfiga}
  \end{subfigure}
 \begin{subfigure}{0.4\linewidth}
    \centering
    \includegraphics[width=\linewidth]{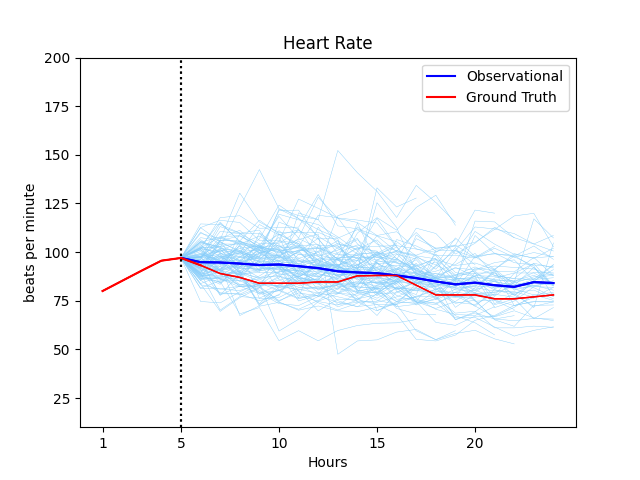}
    \caption{HR}
    \label{fig:subfigb}
  \end{subfigure}
  \caption{\small{Illustration of predictive check on an example MIMIC test set patient. G-Transformer simulated systolic blood pressure (SBP) and heart rate (HR) trajectories (100 Monte Carlo simulations in light blue, average in solid dark blue) compared with ground truth (red) for one patient with predicted treatments under the observational treatment regime  (simulation starting at hour 6).}}
\label{fig:mimic_MC_pt97}
\end{figure}

 Figure \ref{fig:mimic_MC_pt97} illustrates the Monte Carlo simulations generated by G-Transformer for selected variables, specifically systolic blood pressure (SBP) and heart rate (HR), given an example MIMIC patient from test dataset. We observe that the predicted trajectory averaged from 100 Monte Carlo simulations follow closely with the ground-truth trajectory. 


\paragraph{Predictive Check: 24-Hour Outcomes Predictions}

Table \ref{tab:MIMIC_AUC_24H} presents AUCs in predicting clinical outcomes within the first 24 hours of ICU admission at varying simulation start times from G-Transformer and G-Net. There are four outcomes of interest: diagnosis of pulmonary edema, use of a mechanical ventilator (MV), administration of diuretics, and dialysis. We observe that the predicted 24-hour clinical outcomes of interest tend to be more accurate as the value of $k$ increases, and G-Transformer performs slightly better than G-Net when conditioned on a longer history (i.e. $k$=6) except in the case of predicting diuretics use. 

\begin{table*}[h!]
\centering 
\caption{\small{MIMIC-IV Experiments: Predictive check on
clinical outcomes within the first 24 hours of ICU admission at varying simulation start times. Values reported are AUCs. Edema represents presence of Pulmonary Edema.}}
\label{tab:MIMIC_AUC_24H}
\begin{tabular}{ c | c |cccc} 
 \hline 
 \textbf{Model} & k & \textbf{Edema}  & \textbf{MV} & \textbf{Diuretics} & \textbf{Dialysis} \\
 \hline \hline

 \textbf{G-Net} & 2 & 0.802 & \textbf{0.898} & \textbf{0.735} &  0.878  \\

 \textbf{G-Transformer} & 2 & \textbf{0.817} &  0.896  &  0.644 & \textbf{0.880}  \\
 \hline

  \textbf{G-Net} & 6 & 0.850 & 0.938 & \textbf{0.713} & 0.846\\

 \textbf{G-Transformer} & 6 & \textbf{0.864} &  \textbf{0.945}  &  0.650 & \textbf{0.958}  \\
 \hline
 
\end{tabular}
\end{table*}

\subsection{Counterfactual Experiments} 


Our counterfactual strategies are adapted from established clinical trials studying the early treatment of sepsis. This strategy imposes a fluid cap on the total amount of fluids intake of a patient to $X$ liters, and ceases administration once the fluid cap is reached or when the patient is fluid-overloaded. 
More specifically, this strategy was based on the Crystalloid Liberal or Vasopressors Early Resuscitation in Sepsis (CLOVERS) clinical trial \citep{clovers}. For a patient with blood pressure below 65mmHg at time $t$, a 1000mL bolus was administered if the total volume of fluids (including both treatment and maintenance) they received up until that time point did not exceed $X$ liters and if they did not exhibit any signs of fluid overload (as indicated by the presence of pulmonary edema based on chest x-ray radiology reports). 
We experimented with
imposing a fluid cap of $5$L as fluid conservative strategies, and a fluid strategy with no fluid cap to represent a fluid liberal strategy.

\paragraph{Illustrative Examples in Counterfactual Prediction}

\begin{figure}[h!]
\centering
\includegraphics[width=0.32\linewidth]{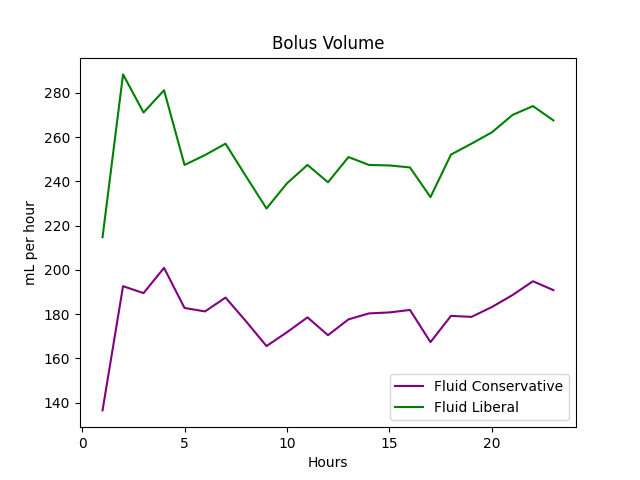}
\includegraphics[width=0.32\linewidth]{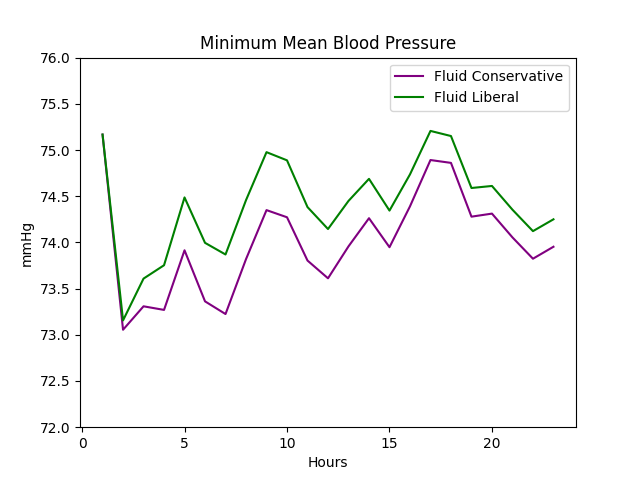}
\includegraphics[width=0.32\linewidth]{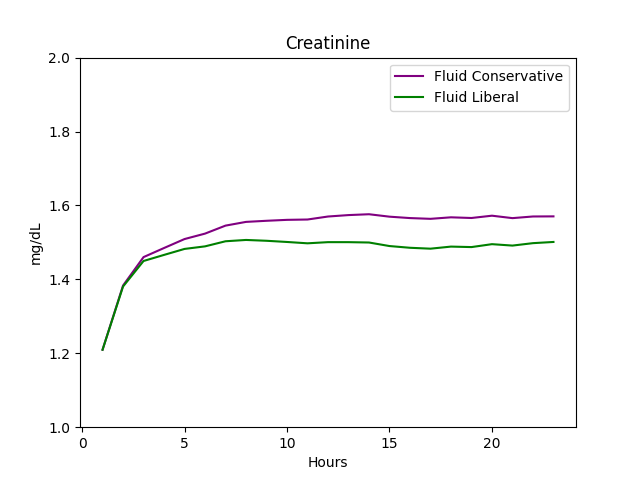}
\caption{\small{Illustrative examples of population-level trajectories of selected covariates under counterfactual fluids strategies. G-Transformer's predictions for selected covariates under various counterfactual fluids strategies, including one fluids conservative (purple) strategy with 5L fluids cap, and a fluids liberal (green) strategy without fluid cap. 
}}
\label{fig:obs_traj_k1_conservative_Liberal}  
\end{figure}

We used G-Transformer to simulate covariate trajectories for patients in the test dataset under one conservative fluids strategies (with 5L fluid cap) and a fluid liberal strategy (with no cap).  Selected covariates are presented in 
Figure \ref{fig:obs_traj_k1_conservative_Liberal}, allowing for comparison of the two counterfactual regimes.  Population-level average trajectories were calculated from the test set (N=894). Counterfactual strategies are applied starting at time step 1, conditioned on observations up until the first hour in the ICU. For each patient, we perform 100 Monte Carlo simulations. The plotted values are averaged across simulated trajectories for all test patients.




We provide illustrative examples of patient trajectories at a population-level to demonstrate G-Transformer's capability to generate clinically meaningful counterfactual predictions. The trends depicted in Figure \ref{fig:obs_traj_k1_conservative_Liberal}
generally align with our physiological and clinical expectations. Under fluid liberal strategies, the bolus volume and blood pressure exhibit higher values, following the expected trends at a population level. Conversely, the higher levels in creatinine   under the fluids conservative regime also align with expected outcomes.

Readers should exercise caution when interpreting these plots in Figure \ref{fig:obs_traj_k1_conservative_Liberal} as indicators of treatment effects, as the number of patients (or trajectories) contributing to the average may not remain constant throughout the entire 24-hour period. Simulated trajectories may end prematurely within the 24-hour period due to death or discharge.

\section{Discussion and Conclusion} 

In this paper, we present G-Transformer, a Transformer-based framework supporting g-computation, to estimate clinical outcomes under counterfactual treatment strategies using both synthetic datasets and a real-world sepsis patient cohort. Our experiments using Cancer Growth and CVSim data, demonstrated that G-Transformer achieved better overall performance compared to other state-of-the-art methods.
Using a real-world sepsis dataset, we demonstrated G-Transformer's potential clinical utility in forecasting covariate trajectories under alternative counterfactual fluid limiting regimes. 
Although we focus on applications in healthcare in this study, the G-Transformer architecture can be applied to other sequential decision making tasks that involve dynamic and time-varying interventions.


\paragraph{Limitations}

One limitation of this work is that counterfactual predictive density estimates in our experiments do not take into account uncertainty about model parameter estimate.  Specifically, given G-Transformer parameters, the distribution of the Monte Carlo simulations produced by g-computation algorithm  constitute an estimate of uncertainty about a counterfactual prediction. However, this estimate ignores uncertainty
about the G-Transformer parameter estimates themselves.  An important area of future work for G-Transformer is adding support for quantification of model uncertainty.


\section{Acknowledgements}
The authors are grateful for Professor Roger Mark for his insightful comments, and Stephanie Hu for her contribution in developing an earlier version of the sepsis cohort. L Lehman was in part funded by MIT-IBM Watson AI Lab, and NIH grant R01EB030362.

\bibliography{paper}

\newpage
\appendix

\section{Additional CVSim Results}
\subsection{Alternative G-Transformer Architecture Experiment}
\label{app:1a}

Our proposed G-Transformer training process differs from the traditional Transformer,
in the way it is iteratively trained on time-series data. Conditioning on observations from the first time step, we treat each time step starting from the second one as an independent subsequence, and use this subsequence to predict the next time step. Then, we concatenated the results of the last time step of each sub-sequence to form a complete prediction sequence for optimization. In contrast, traditional Transformers directly use a sequence-to-sequence framework, shifting the training sequence to obtain the sequence that needs to be predicted. The reason for doing this is to meet the needs of g-computation and dynamic prediction. 
While g-computation calculates probability distributions based on the sequence leading up to the time step being predicted, the Transformer is a typical end-to-end model.
We compared the performance of models implemented with these two training methods on the CVSim dataset, and the results are shown in Table \ref{tab:cvsim_individual_RMSE_gc1_gt_architecture}. 

We can observe that models trained using the sequence-to-sequence approach, including the sequence-to-sequence G-Transformer (Seq2seq G-Transformer) and Causal Transformer, perform notably worse in term of individual-level RMSE compared to proposed G-Transformer in this paper, which uses the iterative training approach. One possible explanation is that during the simulation process, we need to dynamically model the outcomes and treatment variables. This simulation algorithm and our training algorithm share a similar step-by-step approach in term of generating results. Therefore, although both training methods attempt to optimize MSE loss over sequences of the same time step length, our iterative training approach can capture some biases introduced when using predicted outcomes and treatment variables as input for the prediction of next time step, thereby achieving better predictive performance. 

\begin{figure}[h]
\centering
\includegraphics[width=0.5\textwidth]{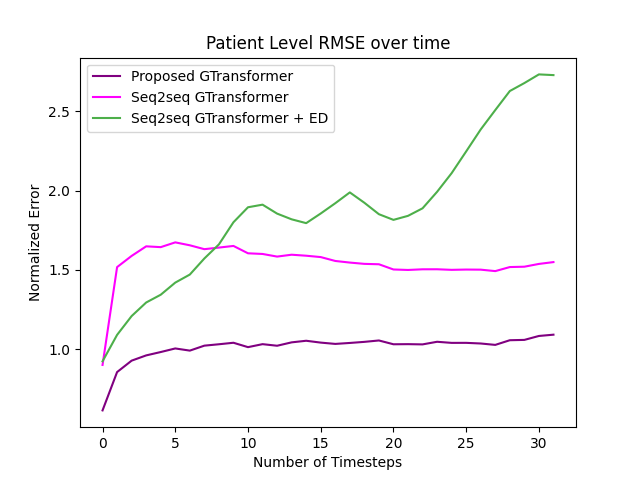}
\caption{\small{Individual-level RMSE over time under \textit{g$_{c1}$} from different variants of G-Transformer}}
\label{fig:seq2seq_rmse_plot}
\end{figure}

Furthermore, as demonstrated in Table \ref{tab:cvsim_individual_RMSE_gc1_gt_architecture} and Figure \ref{fig:seq2seq_rmse_plot}, our experiments with the full Transformer architecture variant, the sequence-to-sequence G-Transformer featuring the full Encoder-Decoder architecture (Seq2seq G-Transformer + ED), did not perform as expected, possibly due to overfitting caused by an excessive number of parameters. Particularly over longer time intervals, the large number of parameters introduced by the decoder may have adversely affected the simulation process.

% ...

\begin{table*}[htb!]
\centering 
\caption{\small{CVSim: Counterfactual prediction under \textit{g$_{c1}$}. Performance of different variants of G-Transformer in predicting remaining 32-hour patient covariate trajectories based on first 34-hour patient covariate trajectories. Values reported represent the individual-level RMSE between the predicted and counterfactual trajectories. ED stands for using entire Transformer Encoder-Decoder rather than just using Transformer Encoder.}}

\label{tab:cvsim_individual_RMSE_gc1_gt_architecture}
\begin{tabular}{ c|cc|cc } 
 \hline 
  & Individual Level \\
 \textbf{Model} & \textbf{RMSE$_{gc1}$} \\
 \hline \hline
 
 \textbf{Seq2seq G-Transformer} & 1.550 \\
 \textbf{Seq2seq G-Transformer + ED} & 1.935 \\
 \textbf{Causal Transformer} & 1.473\\
 \textbf{Proposed G-Transformer} & \textbf{1.016} \\
 \hline
\end{tabular}
\end{table*}

\subsection{Individual-Level RMSE Over Time}
\label{appendix:cvsim_ts_rmse}

Table \ref{cvsim_individual_level_rmse_timestamp} presents the selected time step-specific individual-level RMSE across different models in CVSim experiment. G-Transformer generally performs better than linear implementation of g-computation and G-Net across the time steps. 

\begin{table}[h]
 \caption{\small{CVSim: Counterfactual prediction under \textit{g$_{c1}$} and \textit{g$_{c2}$}. Performance of various models in predicting remaining 32-hour patient covariate trajectories based on first 34-hour patient covariate trajectories per time step. Values reported represent individual-level RMSE between the predicted and counterfactual trajectories.}} 
\label{cvsim_individual_level_rmse_timestamp}
    \centering
    \begin{tabular}{c|l|ccc} 
    \toprule
        {} & {}  & Linear & GNet & GT \\
        {} &  $t$  & \tiny{g-comp} & \tiny{} &  \tiny{G-Transformer} \\
        \midrule
        \textit{g$_{c1}$}  & \small{1} & 0.688 & 0.654 & \textbf{0.615}  \\
        {} & \small{8} & 1.062 & \textbf{1.020} & 1.023  \\
        {} & \small{16} & 1.125 & 1.049 & \textbf{1.042} \\
        {} & \small{24} & 1.168 & 1.052 & \textbf{1.047} \\
        {} & \small{32} & 1.255 & 1.106 & \textbf{1.092} \\
        \hline
        \textit{g$_{c2}$}  & \small{1} & 0.791 & 0.752 & \textbf{0.720} \\
        {} & \small{8} & 1.122 & 1.079 & \textbf{1.077}  \\
        {} & \small{16} & 1.208 & 1.131 & \textbf{1.123} \\
        {} & \small{24} & 1.286 & 1.188 & \textbf{1.165} \\
        {} & \small{32} & 1.348 & 1.213 & \textbf{1.186} \\
\bottomrule
    \end{tabular}
\end{table}




\section{MIMIC Experiments}

\subsection{Adjustment of Individual-Level RMSE}
\label{appendix:mimic_corrections}

Individual-level RMSE is adjusted to account for error due to over- and under-prediction in the trajectory length as a result of death and discharge.  We use the same technique as in  \cite{sucounterfactual} to calculate the individual-level RMSE. 
Specifically, to account for potential over or under-predictions in trajectory length compared to ground truth, we adopt the following approach to modify the individual-level RMSE calculation. First, for patients who died within the first 24 hours, we replace all subsequent time steps of continuous variables with zero (normalized) until the 24th hour. Covariates with potential negative values are filled with the normalized minimum value in the dataset, while covariates following a log-normal distribution are filled with a normalized logarithmic zero. Second, for patients who were released from the hospital before the end of the first 24 hours in the ICUs, we fill all time steps following hospital release with the population-mean for all continuous variables (up until hour 24). Third, we do the same for the predicted chains (after the predicted death and release outcomes during the first 24 hours) so that all chains have lengths of 24-hours. We then compute the average RMSE across all patients by comparing the predicted trajectory (averaged across 100 Monte Carlo simulations per patient) and the ground-truth trajectory of each patient across all continuous variables.

\section{CVSim Data Generation}
\label{appendix:cvsim_data}

We follow the same procedure as in \cite{GNet2021} in generating the CVSim dataset. Briefly, a CVSim 6-compartment circulatory model takes as inputs 28 variables that together govern a hemodynamic system. It then deterministically simulates forward in time a set of 25 output variables according to a collection of differential equations (parameterized by the input variables) modeling hemodynamics.
Important variables in CVSim  include arterial pressure (AP), central venous pressure (CVP), total blood volume (TBV), and total peripheral resistance (TPR). In real patients, physicians observe AP and CVP and seek to keep them above a clinically safe threshold. They do this by intervening on TBV (through fluid administration) and TPR (through vasopressors).

We defined simulated treatment interventions that were designed to mimic the impact of fluids and vasopressors. These simulated interventions alter the natural course of the simulation by increasing either TBV (in the case of the simulated fluids intervention) or TPR (in the case of the simulated vasopressor intervention). We generated patients by randomly initiating baseline inputs (which we hid from our models to make this a stochastic modeling problem) within plausible physiologic ranges, then using CVSim to simulate covariates forward, intervening according to the relevant treatment strategy at each timestep. 

Under (stochastic) observational treatment strategy $g_o$, the probability of receiving a non-zero vasopressor or fluid dose at a given time increases as MAP and CVP decrease according to a logistic regression function. Given that a dose is non-zero, the exact amount is drawn from a normal distribution with mean inversely proportional to MAP and CVP. Since all drivers of treatment under $g_o$ are observed in our data, the sequential exchangeability assumption holds and g-computation may be validly applied. 

$g_{c1}$ is similar to $g_o$, except it is a deterministic treatment strategy and the functions linking treatment and dose to covariates have different coefficents. Under $g_{c2}$, treatment is always withheld.

\section{MIMIC Data and Cohort}

\label{appendix:mimic_data}


Our cohort was limited to ICU stays in which the patient was identified as septic under the Third International Consensus Definitions for Sepsis and Septic Shock (Sepsis-3) \cite{sepsis3-def}.  Patients with missing records of pre-ICU fluids or admitted to the ICU following cardiac, vascular, or trauma surgery were removed.

We use AUC to evaluate our 24 hour outcome prediction performance. For each delay $k$, we do not include patients who have experienced death or release before hour $k$. A patient is positively labelled if they experience an outcome from hours $k$ to 24. Table \ref{tab:24H_cohort_stats} presents these test cohort statistics.

\begin{table*}[h!]
\small
\centering 
\caption{\small{24-Hour Cohort Statistics. Size of the test sets used to evaluate AUCs and the respective percent of patients experiencing each outcome from hour $k$ to 24.}}
\label{tab:24H_cohort_stats}
\begin{tabular}{ c |cccc} 
 \hline 
 k & \textbf{Edema}  & \textbf{MV} & \textbf{Diuretics} & \textbf{Dialysis}  \\
 \hline
  2& (894, 29.98\%)& (894, 40.94\%)&(894, 15.10\%)&(894, 2.68\%) \\


  6& (894, 29.75\%)& (894, 38.81\%)&(894, 12.75\%)&(894, 2.68\%) \\

\hline

\end{tabular}
 
\end{table*}

As predictors to our model, we selected covariates that are typically monitored in the ICU and important for determining sepsis intervention strategies, as well as potential confounders. The covariates we used were similar to that of Li et al. \cite{GNet2021}, encompassing, but not limited to, basic demographic information, an Elixhauser comorbidity score, a SOFA score, laboratory values and vital signs, and urine output \cite{GNet2021}. A comprehensive list is provided in Tables \ref{tab:static_vars} and \ref{tab:time_varying_vars}. The demographics, comorbidities, and pre-ICU fluids were regarded as static while the remaining variables were modeled and regarded as dynamic (time-varying). 

\begin{table}
\centering 
\caption{MIMIC static variables. All variables were used as inputs to our models. }
\vspace{1em}
\label{tab:static_vars}
\begin{tabular}{ c|c|c } 
 \hline 
 \textbf{Variable Name} & \textbf{Variable Type} & \textbf{Units} \\ 
 \hline \hline 
 Age & Continuous & years \\
 Gender & Binary & N/A \\
 Pre-ICU Fluid Amount & Continuous & mL \\
 Elixhauser Score & Continuous & N/A \\
 End Stage Renal Failure & Binary & N/A \\
 Congestive Heart Failure & Binary & N/A \\
 \hline 
\end{tabular}
\end{table}


\begin{table}
\centering 
\small
\caption{MIMIC time-varying variables. All variables were used as inputs to our models, and boluses and vasopressors were also intervention variables. *Refers to maintenance fluids (not an intervention).}
\vspace{1em}
\label{tab:time_varying_vars}
\begin{tabular}{ c|c|c } 
 \hline 
 \textbf{Variable Name} & \textbf{Variable Type} & \textbf{Units} \\ 
 \hline \hline 
 Heart Rate & Continuous & beats/min \\
 Diastolic Blood Pressure & Continuous & mmHg \\
 Systolic Blood Pressure & Continuous & mmHg \\
 Mean Blood Pressure & Continuous & mmHg \\
 Minimum Mean Blood Pressure & Continuous & mmHg \\
 Minimum Change in Mean Blood Pressure from Baseline & Continuous & mmHg\\
 Minimum Mean Blood Pressure from Baseline & Continuous & mmHg\\
 Minimum Change in Mean Blood Pressure from Previous & Continuous & mmHg\\
 Temperature & Continuous & degree C \\
 SOFA Score & Treated as Continuous & N/A \\
 Change in SOFA Score from Baseline & Treated as Continuous & N/A\\
 Change in SOFA Score from Previous & Treated as Continuous & N/A\\
 Platelet & Continuous & counts/$10^9$L \\
 Hemoglobin & Continuous & g/dL \\
 Calcium & Continuous & mg/dL \\
 BUN & Continuous & mmol/L \\
 Creatinine & Continuous & mg/dL \\
 Bicarbonate & Continuous & mmol/L \\
 Lactate & Continuous & mmol/L \\
 O2 Requirement Level & Continuous & N/A\\
 Change in O2 from Baseline & Continuous & N/A\\
 Change in O2 from Previous & Continuous & N/A\\
 pO2 & Continuous & mmHg \\
 sO2 & Continuous & \% \\
 spO2 & Continuous & \% \\
 pCO2 & Continuous & mmHg \\
 Total CO2 & Continuous & mEq/L \\
 pH & Continuous & Numerical[1,14] \\
 Base excess & Continuous & mmol/L \\
 Weight & Continuous & kgs \\
 Change in Weight & Continuous & kgs\\
 Respiratory Rate & Continuous & breaths/min \\
 Fluid Volume* & Continuous & mL \\
 Urine Output & Continuous & mL \\
 Cumulative Edema & Binary & N/A\\
 Pulmonary Edema Indicator & Binary & N/A \\
 Diuretics Indicator & Binary & N/A \\
 Dialysis Indicator & Binary & N/A \\
 Mechanical Ventilation Indicator & Binary & N/A \\
 Vasopressor Indicator & Binary & N/A \\
 Bolus Volume & Continuous & mL \\
 In-Hospital Mortality Indicator & Binary & N/A \\
 Release Indicator & Binary & N/A \\
 Cumulative Fluids & Continuous & mL\\
 Fluid Balance in ICU & Continuous & mL\\
 \hline 
\end{tabular}
\vspace{1em}
\justifying
\end{table}



\section{Hyperparameter Settings}

\subsection{Settings in Cancer Growth Experiments}

Hyperparameter optimization on Cancer Growth data was performed by searching over the hyperparameter space shown in \textbf{Table \ref{tab:cancer_hyperparams}} with optimal parameters starred. 
The experiments were performed on NVIDIA Tesla K80 GPU with 12 vCPUs + 110 GB memory. Hyperparameter settings of rMSN and CRN were as reported in \cite{GNet2021}.

\begin{table}[h!]
    \small 
    \caption{Hyperparameter search space in Cancer Growth experiments. }
    \begin{center}
    \begin{tabular}{c|cc}
    \hline \\
    {} & \textbf{Hyperparameters} &\textbf{Search Range} \\
    \hline \hline \\
    \textbf{Linear (g-comp)} & Learning Rate & 0.001$^*$, 0.01 \\
    \hline \\
    {} & Number of Layers & {1$^*$}, 2 \\
    {} & Hidden Dimension (Continuous) & 8, 16, 32, {64$^*$} \\
    \textbf{G-Net} & Learning Rate & 0.001, {0.01$^*$} \\
    \hline \\
    {} & Hidden Dimension (Continuous) & 4, 16, {64$^*$} \\
    {} & Number of Layers & 2, {4$^*$}, 6 \\
    {} & Hidden Dimension & 32, {64$^*$}, 128 \\
    {} & Batch Size & {16$^*$}, 32 \\
    \textbf{G-Transformer} & Learning Rate & 0.0001 \\
    
    \hline 
\end{tabular}
\end{center}
    \label{tab:cancer_hyperparams}
\end{table}

\subsection{Settings in CVSim Data}

Hyperparameter optimization on CVSim data was performed by searching over the hyperparameter space shown in \textbf{Table \ref{tab:cvsim_hyperparams}} with optimal parameters starred. All models were trained using the Adam optimizer with early stopping with patience of 10 epochs for a maximum of 50 epochs. Also, we utilized the \texttt{CosineAnnealingWarmRestarts} scheduler from PyTorch's optimization module, configured with parameters $T_0 = 10$ and $\text{eta\_min} = 0.00001$, to adjust the learning rate following a cosine annealing pattern.
The experiments were performed on NVIDIA Tesla T4 GPU with 8 vCPUs + 15GB of dedicated memory.

\begin{table}[h!]
    \small 
    \caption{Hyperparameter search space in CVSim experiments. }
    \begin{center}
    \begin{tabular}{c|cc}
    \hline \\
    {} & \textbf{Hyperparameters} &\textbf{Search Range} \\
    \hline \hline \\
    {} & Batch Size & 16 \\
    \textbf{Linear (g-comp)} & Learning Rate & 0.0001 \\
    \hline \\
    {} & Number of Layers  & {2$^*$}, 3\\
    {} & Hidden Dimension (Categorical) & 64, {128$^*$} \\
    {} & Hidden Dimension (Continuous) & 64, {128$^*$} \\
    {} & Batch Size & 16 \\
    \textbf{G-Net} & Learning Rate & 0.0001 \\
    \hline \\
    {} & Number of Layers & 3 \\
    {} & Hidden Dimension & 32, 64, {128$^*$} \\
    {} & Batch Size & 16 \\
    \textbf{G-Transformer} & Learning Rate & 0.0001 \\
    
    \hline 
\end{tabular}
\end{center}
    \label{tab:cvsim_hyperparams}
\end{table}

\subsection{Settings in MIMIC Data}

Hyperparameter optimization on MIMIC data was performed by searching over the hyperparameter space shown in \textbf{Table \ref{tab:mimic_hyperparams}} with optimal parameters starred. All models were trained using the Adam optimizer with early stopping with patience of 10 epochs for a maximum of 50 epochs. For each covariate, the corresponding Transformer encoder in G-Transformer and LSTM in G-Net were optimized through a grid search of the following hyperparameters. The experiments were performed on NVIDIA Tesla T4 GPU with 8 vCPUs + 15GB of dedicated memory.

\begin{table}[h!]
    \small 
    \caption{Hyperparameter search space in MIMIC experiments. }
    \begin{center}
    \begin{tabular}{c|cc}
    \hline \\
    {} & \textbf{Hyperparameters} &\textbf{Search Range} \\
    \hline \hline \\
    {} & Number of Layers  & 2\\
    {} & Hidden Dimension & 64 \\
    {} & Batch Size & 32 \\
    {} & Weight Decay & 0.001 \\
    \textbf{G-Net} & Learning Rate & 0.001, 0.0001, 0.00001 \\
    \hline \\
    {} & Number of Layers & 2 \\
    {} & Hidden Dimension & 64 \\
    {} & Batch Size & 32 \\
    {} & Weight Decay & 0.001 \\
    \textbf{G-Transformer} & Learning Rate & 0.001, 0.0001, 0.00001 \\
    
    \hline 
\end{tabular}
\end{center}
    \label{tab:mimic_hyperparams}
\end{table}

\subsection{Settings for Causal Transformer}
Different from the Causal Transformer paper, we did not generate different counterfactual trajectories on the Cancer Growth dataset at each time step using \textit{Single sliding treatment} or \textit{Random trajectories}. For each patient in the test set, we inserted our counterfactual treatment at the last 5 time steps and applied this counterfactual treatment in each subsequent time step. In addition, for multi-step time prediction with Causal Transformer, we predicted the next 4 time steps and $\tau=5$ in this dataset. 
In both the CVSim dataset and the MIMIC dataset, since the experiment setting does not allow access to the actual treatment, we modified parts of the Causal Transformer's code to enable the model to use the predicted treatment as input for predicting the outcomes of the next time step. Specifically, we set the $\tau$ of Causal Transformer to 1, and repeat the entire prediction process $n$ times. Between each prediction step, we use a function to calculate whether the treatment is needed. The obtained treatment and the outcome predicted are then added as inputs to supplement the end of the input queue.

\textbf{Hyperparameter Settings} For MIMIC and Cancer growth dataset, given that the Causal Transformer has already provided rich information on hyperparameter settings in the original paper \citep{melnychuk2022causal}, we used their hyperparameter settings in experiments. For CVSim dataset, we made slight adjustments to the hidden dimension for experimentation as its format is similar to that of the MIMIC dataset. The specific hyperparameter settings are shown in the table \ref{tab:CT_hyperparams}.

\begin{table}[h!]
    \small 
    \caption{Hyperparameter setting for Causal Transformer in different experiments. }
    \begin{center}
    \begin{tabular}{c|cc}
    \hline \\
    {} & \textbf{Hyperparameters} &\textbf{Search Range} \\
    \hline \hline \\
    {} & Dropout rates &  0.2 \\
    {} & Number of Layers &  2 \\
    {} & Number of attention heads  & 3 \\
    {} & Number of epochs &  100 \\
    {} & Hidden Dimension &  44 \\
    \textbf{MIMIC} & Learning Rate & 0.001  \\
    \hline \\
    {} & Dropout rates &  0.1 \\
    {} & Number of Layers &  2 \\
    {} & Number of attention heads  & 3 \\
    {} & Number of epochs &  150 \\
    {} & Hidden Dimension &  48 \\
    \textbf{Cancer Growth} & Learning Rate & 0.0001  \\
    \hline \\
    {} & Dropout rates &  0.2 \\
    {} & Number of Layers &  2 \\
    {} & Number of attention heads  & 3 \\
    {} & Number of epochs &  100 \\
    {} & Hidden Dimension &  24 \\
    \textbf{CVSim} & Learning Rate & 0.0001  \\
    \hline 

\end{tabular}
\end{center}
    \label{tab:CT_hyperparams}
\end{table}

\end{document}